\documentclass[sigplan,screen]{acmart}
\renewcommand\footnotetextcopyrightpermission[1]{} % removes footnote with conference information in first column
\AtBeginDocument{%
  \providecommand\BibTeX{{%
    \normalfont B\kern-0.5em{\scshape i\kern-0.25em b}\kern-0.8em\TeX}}}

\setcopyright{None}
\copyrightyear{2019}
\acmYear{2019}
\acmDOI{}

\acmConference[]
\acmBooktitle{}

\begin{document}

\title{Explainable AI \\Reinforcement Learning Agents for\\Residential Cost Savings}


\author{Hareesh Kumar}
\affiliation{
   \institution{CSE, IIT Bombay}
}
 \email{hareesh@iitb.ac.in}

\author{Priyanka Mary Mammen}
\affiliation{
   \institution{CSE, IIT Bombay}
}
 \email{priyankam@cse.iitb.ac.in}

\author{Krithi Ramamritham}
\affiliation{
   \institution{CSE, IIT Bombay}
}
 \email{krithi@iitb.ac.in}

% \author{ABC}
% \affiliation{%

\renewcommand{\shortauthors}{Hareesh, et al.}

\begin{abstract}
%Motivated by the recent advancements in the Deep Reinforcement Learning (RL), we have developed an RL agent to manage the operation of storage devices in a household and is designed to maximize demand-side cost savings. The proposed technique is data-driven, and the RL agent learns from scratch how to efficiently use the energy storage device given variable tariff-structures. In most of the studies, the RL agent is considered as a black box, and how the agent has learned is often ignored. We explain the learning progression of the RL agent, and the strategies it follows based on the capacity of the storage device. 

% Ch
Motivated by recent advancements in deep Reinforcement Learning (RL), we have developed an RL agent to manage the operation of storage devices in a household and is designed to maximize demand-side cost savings. The proposed technique is data-driven, and the RL agent learns from scratch how to efficiently use the energy storage device given variable tariff structures. In most of the studies, the RL agent is considered as a black box, and how the agent has learned is often ignored. We explain the learning progression of the RL agent, and the strategies it follows based on the capacity of the storage device. 
\end{abstract}



\maketitle

\section{Introduction}

 Renewable sources of energy account for more than half of the energy supply sources today and are set to penetrate global energy markets at a rapid pace \cite{bp}. But, quite often, there is a mismatch between the time of  renewable energy generation and the actual time when there is a demand for energy. Renewable energy generators like Solar Photo-voltaic Cells have a high potential for energy generation during the noon hours, but the  demand for energy in residential buildings is higher during the evening hours as most of the people spend the day hours at work.
This imbalance in demand and supply in modern (smart) power grids has forced utilities to come up with variable pricing methods such as Time of Day (ToD) or Dynamic Pricing structures to motivate consumers to consume energy during times of higher availability.
The price of electricity is kept lower during hours of relatively higher power supply and higher tariffs are levied for high demand hours. The utilities indirectly reward  users who consume during high supply hours and penalize the consumers who consume during high demand hours. Some utilities also directly pay consumers to reduce energy consumption during high demand hours. Such Demand Response(DR) occasions call for intelligent decision making by the entities involved, helping reduce energy costs for both consumers and producers.

In some developing countries, when there is high demand but less supply,  utilities adopt rolling blackouts where specific regions  suffer from blackouts for a few hours \cite{beg2013integrating}.  Frequent blackouts have led to an increase in the number of households that purchase battery-based energy storage devices for backup power supply when energy from the grid is not available.
%Over the years, reducing costs and increased efficiency of these batteries \cite{zhang2018energy} have led to an increase in their adoption.
%Ch 
Over the years, reducing costs and increased efficiency of these batteries \cite{zhang2018energy} have  led to an increase in their adoption.

%Talk about the problem here
 Demand Side Energy Management\cite{beaudin2015home, strbac2008demand, sundstrom2010optimization} offers a set of techniques to improve building energy consumption via load shifting or load reduction. However, the deferrable loads are limited to dishwashers, washing machine, air-conditioners, plug-in electric vehicles, etc. With a significant fraction of non-deferrable loads, it is easier to meet the energy needs using energy storage systems, under variable energy pricing. Modeling an efficient Energy Management System which can optimize the usage of battery can lead to cost savings for the consumers. This could also help in increasing the stability of the grid by decreasing the demands during energy deficit hours. Traditional methods rely on rule-based techniques to optimize battery operation\cite{tant2012multiobjective, arcos2016fuzzy}. But, these methods need to be flexible
 and must adapt to changes in the tariff methods.

%talk about advances in the RL and how it can be useful
Reinforcement Learning(RL) has gathered a lot of popularity especially in the gaming domain, where it has been shown that it can surpass  human-level decision-making in most of the games\cite{silver2017mastering, mnih2015human1}. In this paper, we model a deep reinforcement learning agent which can learn to work under a variable pricing regime to provide cost savings to the consumers of energy. The goal of the RL agent is to maximize cost savings, i.e., to reduce the cost that has to be paid to the utility. The cost savings of the agent is calculated using the formula below
\begin{equation}
    cost\_saving = \bigg(  1 - \frac {Cost(RL Agent)}{Cost(BaseLine)} \bigg) *100
\end{equation}{}
where $Cost(BaseLine)$ is the baseline cost, i.e., the cost that the consumer would have paid in the absence of the RL agent, $Cost(RL Agent)$ is the cost incurred when incorporating the trained RL agent %Ch $Cost(RL Agent)$ is the cost incurred when incorporating the trained RL agent

% you can integrate related work here also.
\section{Related Work}

Extensive studies have been carried out on the optimal control of energy storage systems. Many of them \cite{babacan2017distributed, ratnam2015optimization,kazhamiaka2017influence}try to formulate it as an optimization problem where the decisions are taken beforehand. For instance, in \cite{kazhamiaka2017influence} a linear integer program is solved to determine the battery operation to improve the profitability over a period of 20 years. Moreover, these papers do not consider the uncertainty in consumption and also rely on accurate prediction and information. Few papers \cite{guan2015reinforcement} that consider the uncertainty in information cannot handle problems with a large state space.  Advances in RL offer an opportunity to solve problems with large state space. Also, when compared to traditional rule-base approaches, RL algorithms learn the best control policy by itself. Works such as \cite{sekizaki2015intelligent, shi2017echo,berlink2015intelligent, qiu2015heterogeneous} try to solve problems in energy domain with new RL techniques. However, many of them are formulated for the purpose of better battery utilization along with renewables and present the agent as a black box \cite{berlink2015intelligent, qiu2015heterogeneous}. Although solar PV deployments are not common in households, most of them adopt battery storage systems for managing energy shortage situations. None of the current RL approaches propose an intelligent decision-making agent for battery storage operations (excluding renewables) in households under various electricity tariff structures.

\section{Modeling the Energy Management System}
\begin{figure}[H]
 \includegraphics[totalheight=4cm]{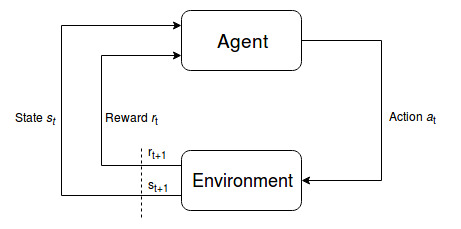}
 \caption{Reinforcement Learning} 
 \label{img:rl_intro}
\end{figure}

Reinforcement Learning is the study of decision making over time in complicated environments. Fig \ref{img:rl_intro} shows the RL agent, at time $t$, the agent is in  state $s_t$, performs an action $a_t$ from the list of possible actions which is executed in the Environment, and the reward $r_t$ for taking that action along with the next state $s_{t+1}$ is returned to the agent. 
In our case, the agent is the controller box that performs actions, and the environment is the utility. State $s_t$ is the environment's private representation, action $a_t$ is the combination of charge/discharge operations that the agent performs. The rewards are computed based on how good the chosen action is. The rewards are computed by the Environment and are returned to the Agent. 

\subsection{Observation State}
Observation state is represented as a continuous array which contains the Time of the Day Pricing or the forecasted price for the next $n$ hours along with the status of the battery and the current load. Other data points (if any) could also be added to this list. 
\begin{equation}
    s_t = [ t_1, t_2, t_3 .... t_n, battery\_status, current\_load, o_1, o_2 ..  o_n ]
\end{equation}

where $t_1, t_2, t_3 .... t_n$ are the Time of the Day Pricing or the forecasted price for $n$ hours, and  $battery\_status$ is the status of the battery,   $current\_load$ is the current demand and $o_1, o_2 .. o_n$ represent any other information which could be appended. The observation state is not labeled, the information about what the values in the list signify is not known to the agent.

\subsection{Action Space} 
Action Space is the set of possible actions that the agent can perform in a State. There are three possible actions that the Reinforcement  Agent can choose.
\begin{itemize}
    \item Grid: The Smart Grid fulfills the  consumer demand for energy
    \item Battery Discharge+Grid: The Energy Storage device fulfills the maximum possible consumer demand, and the Smart Grid fulfills the rest of the consumer demand. 
    \item Battery Charge + Grid: The Smart Grid fulfills the total consumer demand, as well as the energy storage device demand.
    
\end{itemize}

\subsection{Reward Function}
The goal of the Reinforcement Learning agent is to maximize the expected cumulative reward. 
\begin{equation}
    R = \sum_{t=0}^\infty \gamma^t r_t
\end{equation}
where $\gamma$ with value between 0 and 1 is  the discount factor; the larger the value of $\gamma$  the more importance is given to the future reward and $r_t$ is the reward at time $t$.

\textbf{Episodic Reward:}  In this case, we have a starting and an ending point, and the reward is computed and returned to the agent at the ending point. In the case of episodic reward the reward could be given as the negative of the total cost that the user will have to pay the utility.

\textbf{Continuous Reward:} The continuous reward is an immediate reward given for every action that the agent performs. E.g., let us assume we are training a robot to walk, we can give rewards in the form of linear/exponential function when it balances itself and walks. If the robot crashes then we can give a high negative reward and terminate.

Here we model the reward as a continuous reward as it intuitively feels that the agent can learn better if continuous reward is given. The reward at time \textit{t} is computed as the cost to be paid to the utility to consume from the grid
\begin{equation}\label{eq:reward}
R(t) = -Cost\_electricity(t) \times demand(t) + penalty(demand(t))
\end{equation}{}
where $R(t)$ is the reward at time $t$.
%In some of the cases, there is a penalty that has to be paid for exceeding the maximum demand limit, calculated by penalty(x).
%Ch 
In some cases, there is a penalty to be paid for exceeding the maximum demand limit, calculated by penalty(x).

The incentives in the case of Demand Response can also be integrated with the penalty function. Negation is used here as we want the RL agent to minimize the cost. 
%The function used to calculate the rewards, incentives or the real-time pricing is not known to the agent, but the result is known i.e, the agent will know the price at different times but does not know \textit{how} the values are computed.
%Ch 
The function used to calculate the rewards, incentives or the real-time pricing is not known to the agent, but the result is known i.e, the agent will know the price at different times but does not know \textit{how} the values are computed.

\begin{figure*}
 \includegraphics[totalheight=6cm]{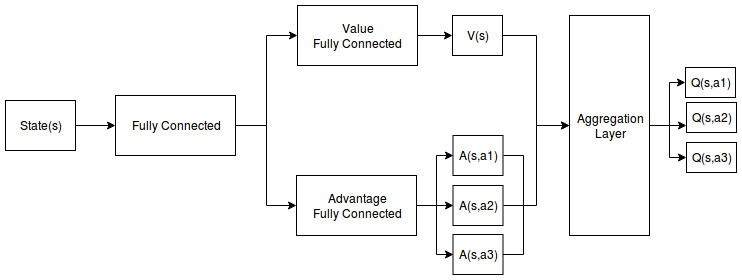}
 \caption{Double Dueling Deep Q-Learning Network (DDDQN)} 
 \label{img:ddqn}
\end{figure*}

\section{Framework for the RL agent}
We use Q-Learning to learn a policy which will help the RL agent to perform optimal action given a state. Given a state $s$ and an action $a$, $Q(s,a)$ denotes how good/bad it is to take action $a$ being in state $s$. Q-Value of a state is computed using the Bellman Equation, Eq. \ref{eq:q-value}, by updating the Q-Values until convergence.\\

\begin{equation} \label{eq:q-value}
    Q^{new}(s_t,a_t) = (1-\alpha)*Q(s_t,a_t) + \alpha *(r_{t} + \gamma *max_{a} Q(s_{t+1},a))
\end{equation}{}
where $\alpha$ is the learning rate, $Q^{new}(s_t,a_t)$ is the new value for state 
$Q(s_t,a_t)$, 
$\gamma$ is the discount factor and 
$max_{a} Q(s_{t+1},a)$ is the maximum possible reward given $s'$ and all possible actions at that state.

\textbf{Deep Q-Learning:} 
Here we use Deep Neural Network to approximate the Q-Values, which would have been obtained by Eq. \ref{eq:q-value}. Neural Network takes the state $s$ as input and outputs the Q-Value for every action that could be taken. The agent chooses the action which results in the maximum Q-Value at each step. Rectified Linear unit(ReLU) is used as an activation function for the hidden layers in the neural network.The target $Q^*(s_t, a_t)$ values in the neural network is estimated using equation  
\begin{equation}
    Q^* (s_t,a_t) = r_{t+1}+ \gamma max_{a_t+1} - Q(s_{t+1}, a_{t+1})
\end{equation}

The loss function for the network is computed as the Mean Squared Error(MSE) of the computed Q-Value of the neural network to the target Q-value. Mini-batch gradient descent method is used to update the parameters in the neural network $\omega = \alpha \triangle\omega$
where $\alpha$ is the learning rate, and the $\triangle\omega$ is the derivative of the loss function $\triangle \omega = \frac{\partial L}{ \partial w} $.\\
The loss function is calculated as
\begin{equation}
L = \frac{1}{2n} \sum_1^{n} [(Q^* (s_t,a_t) - Q(s_t, a_t))^2]
\end{equation}

\subsection{Prioritized Experience Replay(PER)}
 We use Prioritized Experience Replay(PER) \cite{schaul2015prioritized} where the experiences(state, action, reward, next\_state) are stored and chosen based on priorities. %Ch  We use Prioritized Experience Replay(PER) \cite{schaul2015prioritized} where the experiences(state, action, reward, next\_state) are stored and chosen based on priorities.
 The priority of the experience is set based on the predicted value and the target value; higher the difference higher the priority. Probability of being chosen for a reply is given by stochastic prioritization 

\begin{equation}
    P(i) = \frac{p_i^a} {\sum_k p_k^a}
\end{equation}
where $p(i)$ is the priority value of $i$ being selected, hyperparameter $a$ is used to introduce randomness in the experience selection for replay buffer. If $a$=0, then it is pure randomness, if a=1 then select only the experiences with highest priorities. Updates to the network are weighted with Importance Sampling weights(IS), to account for the change in the distribution.The IS weights are updated by reducing the weights of the often seen samples.
\begin{equation}
IS = \Big( \frac {1} {N} * \frac {1} {P(i)} \Big)^b
\end{equation}{}
where N is the size of the Replay buffer size, P(i) is the sampling probability and $b$  controls how the sampling weights affect the learning process. If $b = 0 $ then no importance sampling is applied, and if $b =  1$ then full importance sampling is applied.

\subsection{Fixed Q-targets}
We use the idea of fixed Q-targets introduced by Mnih et al. \cite{mnih2015human1}, by using a separate network with a fixed parameter for estimating target value and at every T step, we copy the parameters from DQN (Deep Q-Network) network to update the target network. This improves the stability of the Neural Network as updates are not made to the target function after every batch of learning.

\begin{equation}
    \triangle w = \alpha[(R + \gamma max_a \hat{Q}(s',a,w^-)) - \hat{Q}(s,a,w)]\triangledown_w \hat{Q}(s,a,w)
\end{equation}{}
where $\triangle w $ is the change in weights, $\alpha$ is the learning rate, 
$\triangledown_w \hat{Q}(s,a,w)$ is the gradient of the predicted Q-value,  $\hat{Q}(s',a,w^-)$ is the current predicted Q-value and $(R + \gamma max_a \hat{Q}(s',a,w^-)) $ is the maximum possible Q-value for the next state predicted from the target network.

\subsection{Double DQN}
Double DQN introduced by Hado van Hasset\cite{van2016deep} is used to handle the problem of  overestimation of the Q-values. The accuracy of the predicted Q-Value depends on the actions that are explored. Taking the maximum Q-value will be noisy during the initial phase of training if non-optimal actions are regularly given higher Q-value than the optimal best action. This also complicates the learning process.
The best action to be taken at the next state (the action with the highest q-value) is computed from the DQN network, and the target Q-Value of taking that action at the next state is computed from the target network.

\begin{equation}
    Q(s,a) = r(s,a) + \gamma Q(s',argmax_a Q(s',a))
\end{equation}
 $Q(s,a)$ is the Q-target, $r(s,a)$ is the reward of taking that action at that state and $\gamma Q(s',argmax_a Q(s',a)$ is the discounted max q value among all possible actions from the next state.

\subsection{Dueling Double DQN}
In a Dueling Double Deep Q- Learning Neural Network (DDDQN) \cite{wang2015dueling}, the value of  Q(s, a) is computed as the sum of the value of being in that state $V(s)$ and the advantage of taking action at that state $A(s, a)$. By decoupling the estimation, intuitively our DDQN can learn which states are (or are not) valuable without having to learn the effect of each action at each state since it's also calculating V(s).Hence, the local minima is not inadvertently chosen (or avoided) since the advantage of taking the action is also considered.

\begin{equation}\label{eq:dddqn}
    Q(s,a,\theta;\alpha,\beta) = V(s;\theta,\beta) + (A(s,a;\theta,\alpha) - \frac{1}{A} \sum_{a'}  A(s,a';\theta,\alpha))
\end{equation}{}

$\theta$ is the common network parameters, $\alpha$ advantage stream parameters, $\beta$ value stream parameters, $\frac{1}{A} \sum_{a'}  A(s,a';\theta,\alpha)$ is the average advantage. This architecture helps boost the training as we can calculate the value of the state without calculating the value for $Q(s, a)$ for each action at that state, it also helps us to find the reliable Q-Values for each action as the value and the advantage are decoupled.

Fig \ref{img:ddqn} shows the DDDQN architecture,.
\textit{Value Fully Connected} is used to calculate the value function of the state and \textit{Advantage Fully Connected} is used to calculate the advantage of taking the action. The aggregation layer performs the Eq. \ref{eq:dddqn}

\section{Experiments and Results}
\begin{figure}[H]
 \hfill \includegraphics[totalheight=5cm]{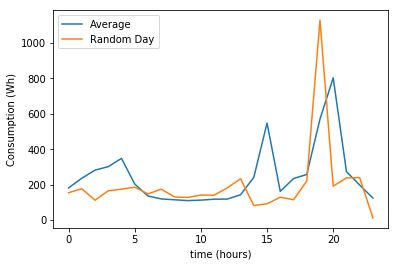} \hspace*{\fill}
 \caption{Sample Consumption} 
 \label{img:sample_consumption}
\end{figure}

 We use the dataset of a high-rise residential building\cite{mammen2018want},  Fig \ref{img:sample_consumption} shows the power consumption of the apartment which was considered under the study. The blue line represents the avg. consumption of the apartment for a month and the yellow line shows the consumption of the apartment for a day.

\subsection{Naive RL agent}
During the training phase, the model was saved for every ten iterations. Then the saved model was evaluated on the test dataset, we then explain the learnings of the agent. %Ch we then explain the learnings of the agent. -- you might want to rephrase this part of the sentence. 

\textbf{Observation Space}
The observation space consists of the Time of Day(ToD) pricing for the next 24 hours along with the battery status and the current load. Time of Day pricing was modeled as shown in Table \ref{tab:tod}
\begin{table}[H]
\begin{tabular}{|l|l|}
 \hline
 ToD - Time Slot & Cost (x/kWh) \\
  \hline
00:00 Hours to 08:00 Hours & 1       \\
08:00 Hours to 16:00 Hours & 3       \\
16:00 Hours to 24:00 Hours & 2       \\
 \hline
\end{tabular}
 \caption{ToD Pricing}
 \label{tab:tod}
\end{table}

The Energy storage device used here is a battery with a capacity of 900Wh and maximum discharge/charge rate set to 300W.\\

\textbf{Rewards}
The rewards at time $t$ are calculated using the Eq \ref{eq:reward} and value for penalty is assumed to be null.

\textbf{Hyper-parameters}
% Grid search was performed to find the hyper-parameters that helps in training the network faster.

\begin{table}[H]
\begin{tabular}{|l|l|}
 \hline
 Hyperparameter & Value \\
  \hline
mini batch size                  & 32       \\
replay memory size              & 10240    \\
discount factor                 & 0.96     \\
learning rate                   & 0.00025  \\
initial exploration             & 1.0      \\
final exploration               & 0.1     \\
 \hline
\end{tabular}
 \caption{Hyper-parameters}
 \label{tab:hyp}
\end{table}
%k justify the numbers

\begin{figure}
\hfill \includegraphics[totalheight=5cm]{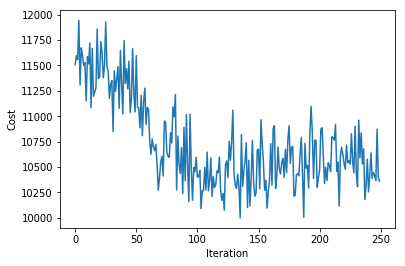} \hspace*{\fill}
 \caption{Battery Status} 
 \label{img:er_cost}
\end{figure}

RL agent was trained on a 30-day residential dataset. The hyper-parameters which were used during training is mentioned in the Table \ref{tab:hyp}.  Fig \ref{img:er_cost} shows the gradient descent of the RL agent, where it tries to reduce the energy cost of the residence.

\subsubsection{Learnings of RL agent}
\begin{figure}[H]
\hfill \includegraphics[totalheight=5cm]{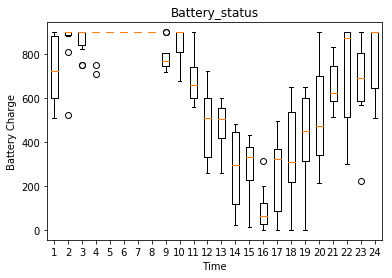} \hspace*{\fill}
 \caption{RL agent learning when to consume from the battery} 
 \label{img:inside_1}
\end{figure}

Around 50 iterations: The agent learns insights on the data given to it and how it affects the gradient ascent. The Agent learns when the cost of the energy is high, and consuming energy from the battery during those hours will help in reducing the energy cost. Fig \ref{img:inside_1} shows agent discharging the battery during the 8-16th hour, which are the hours when the cost of the energy is high.

\begin{figure}[H]
\hfill \includegraphics[totalheight=5cm]{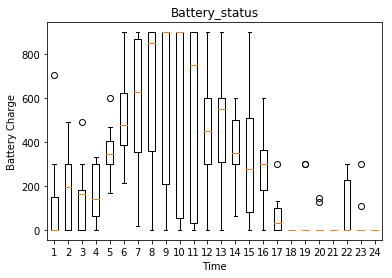} \hspace*{\fill}
 \caption{RL agent learning when to charge the battery} 
 \label{img:inside_2}
\end{figure}

Around 100 iterations: The agent learns when the cost of the energy is cheap, and charging during those hours will help in reducing the energy cost. Fig \ref{img:inside_2} shows agent charging the battery during the 0-8th hour when the cost of energy is low. It should also be observed that the agent sometimes performs random charging and discharging in the slots, and these do not affect the end cost. The agent also learns not to perform any action during the 17-24 hours slot as this does not increase or decrease the cost savings.

\begin{figure}[H]
\hfill \includegraphics[totalheight=5cm]{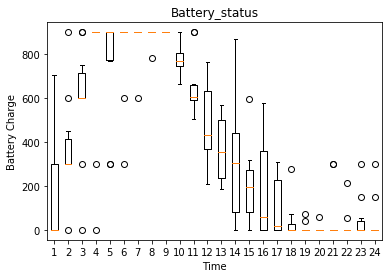} \hspace*{\fill}
 \caption{RL agent learning Time of Day} 
 \label{img:inside_3}
\end{figure}

Around 200 iterations:
Fig \ref{img:inside_2} shows the agent charging the battery as soon as the cost of the energy is low, learns to discharge during the high demand hours, and does not do anything during the other hours.

\subsection{Case Study : ABC City, India}
We model the environment based on Tata Power Tariff, High Tension Residential consumer(Housing Society). We have also included the Time of Day pricing which is not applicable for residential loads but mandatory for most of the consumers in the High Tension load and is optional for a few of the consumers.  High Tension residential consumers are charged at  5Rs per kWh. Table \ref{tab:tod_tata} shows the additional ToD pricing which is followed by Tata Power (Base price of 5Rs has to be added to the ToD specified). The cost of the electricity is lowest from 22.00-06.00 hours with the cost of 4.25Rs/kWh and the cost of the electricity is highest during 18.00-22.00 hours with the cost of 6Rs/kWh.

%Ch The cost of electricity is lowest from 22.00-06.00 hours at 4.25Rs/kWh and highest from 18.00-22.00 hours at 6Rs/kWh.

% \begin{figure}[H]
%  \hfill \includegraphics[totalheight=4cm]{images/ToD_tariff_tata.png} \hspace*{\fill}
%  \caption{Tata ToD Tariff (Base Price 5Rs)} 
%  \label{img:tata_tod}
% \end{figure}

\begin{table}
\begin{tabular}{|l|l|}
 \hline
 ToD - Time Slot & Rs/kWh \\
  \hline
06:00 Hours to 09:00 Hours & 0.00       \\
9:00 Hours to 12:00 Hours & 0.50       \\
12:00 Hours to 18:00 Hours & 0.00       \\
18:00 Hours to 22:00 Hours & 1.00       \\
22:00 Hours to 06:00 Hours & -0.75       \\
 \hline
\end{tabular}
 \caption{Tata ToD Tariff (Base Price 5Rs)}
 \label{tab:tod_tata}
\end{table}

\subsubsection{Hyperparameters}

\begin{table}
\begin{tabular}{|l|l|}
 \hline
 Hyperparameter & Value \\
  \hline
mini batch size                  & 32       \\
replay memory size              & 10240    \\
agent history length            & 15 days  \\
target network update frequency & 5 days   \\
discount factor                 & 0.96     \\
learning rate                   & 0.00025  \\
initial exploration             & 1.0      \\
final exploration               & 0.1     \\
 \hline
\end{tabular}
 \caption{Hyper-parameters}
 \label{tab:hyp_ddqn}
\end{table}

Observation space used here is similar to the one described in the experiment 4.1. The architecture used in the experiments is described in section \ref{img:ddqn}. The hyper-parameters which were used during the training are mentioned in the Table \ref{tab:hyp_ddqn}. The capacity of the battery was varied in the range of 5kWh to 30kWh. The models were initialized with random weights initially. Since deep charge or discharge of the battery reduces the lifetime of the battery, it was ensured that the battery could maximum charge up to 90\% of its total capacity, and the maximum discharge of the battery was limited to 10\% of the total capacity. % Ch Since deep charge or discharge is known to reduce the life of the battery, we ensured that the battery was not charged beyond  90\% of its total capacity, and the maximum discharge rate of the battery was limited to 10\% of the total capacity
The charging capacity and the discharging rate was set to 70\% of the battery's capacity. The loss occurring while charging and discharging was ignored. The cost  and life of the battery was also not taken into consideration. The model was trained on one month's data and tested on the subsequent month's data from the residential dataset.
All the models were trained for 500 epochs.

\subsubsection{Results}

\begin{figure}[H]
 \hfill \includegraphics[totalheight=5cm]{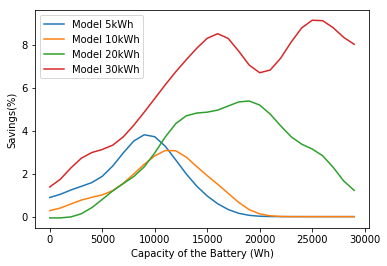} \hspace*{\fill}
 \caption{Storage Capacity vs Cost Savings} 
 \label{img:capacity_saving}
\end{figure}

Fig \ref{img:capacity_saving} shows lower capacity batteries fail to perform when
%well on -> when
%k "on" extra?
tested on high capacity batteries, but the vice-versa is not true. %Ch Fig \ref{img:capacity_saving} shows that lower capacity batteries fail to perform when tested on high capacity batteries, but not vice versa.
This is because the agents trained on the low capacity batteries have not seen states that are experienced by the high capacity batteries, but the high capacity batteries have seen the states that are seen by the low capacity batteries.  
%Ch This is because the agents trained on low capacity batteries have not seen states experienced by high capacity batteries, but high capacity batteries have seen the states that are seen by the low capacity batteries. 

\subsubsection{Low vs High Energy Storage devices}

\begin{figure}[H]
 \hfill \includegraphics[totalheight=5cm]{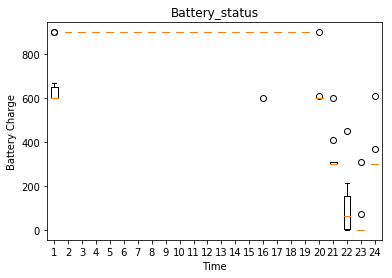} \hspace*{\fill}
 \caption{RL agent on small capacity storage devices} 
 \label{img:ddqn_small_capcity}
\end{figure}

Fig \ref{img:ddqn_small_capcity} shows the common pattern which was observed when the Reinforcement Agent was trained on lower capacity batteries. It also shows that the agent chooses to charge the battery when the cost of energy is low and chooses to discharge when the cost of energy is high. Rest of the time, the agent does not charge/discharge the battery.

\begin{figure}[H]
\hfill \includegraphics[totalheight=5cm]{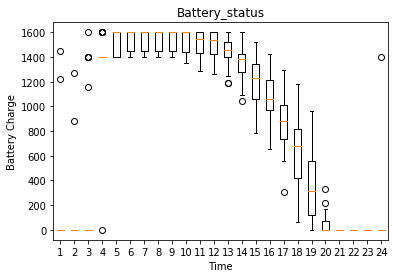} \hspace*{\fill}
 \caption{RL agent on large capacity storage devices}
 \label{img:ddqn_large_capcity}
\end{figure}

Fig \ref{img:ddqn_large_capcity} shows the trained RL agent on high capacity batteries, it can be observed that the agent chooses to completely charge the battery when the cost of energy is low (22.00-06.00 hours) and discharge the battery continuously as soon the cost of energy increases.

\subsubsection{Demand Response}
 In demand response situations, there is a maximum demand limit imposed on the consumers and consuming above this limit results in heavy penalties imposed by the utility.
 We model the Demand Response by setting a maximum demand limit per day as 700Wh for the consumer along with the tariff as mentioned in Table \ref{tab:tod_tata} and the consumer is penalized by adding Rs.2 for every unit exceeding the maximum demand limit.

The Neural Network used for Demand Response was initialized with the results of the earlier models and was fine-tuned to work for demand response.
\begin{figure}[H]
 \hfill \includegraphics[totalheight=5cm]{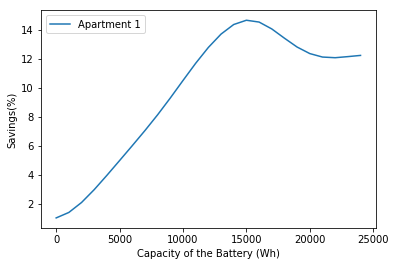} \hspace*{\fill}
 \caption{Demand Response}
 \label{img:rl_dr}
\end{figure}

Fig \ref{img:rl_dr} shows the savings of the agent with ToD pricing and ToD pricing along with the Demand Response Scenario.
 Fig \ref{img:rl_dr} shows an increase in savings when the capacity of the battery is increased. It can be seen that after 15000Wh, the savings tend to flatten around 12\%. Here, it is optimal to use a 15000Wh battery that can achieve 12-14\% cost reduction in this scenario.  
 
%k explain how the right capacity can be identified by taking some points in the graph

Our results show savings of 6-8\% under the ToD tariff method as specified in Table \ref{tab:tod_tata}. Savings of 12-14\% can be obtained if the utility follows the ToD pricing along with rewards from the DR program. The capital cost of the energy storage system, its efficiency, life time of the device, and other factors have to be considered while calculating the payback period.

\section{Conclusion}
This paper presents a deep reinforcement learning based data-driven approach to control an energy storage system. Our results shows that the savings accrued increases when the capacity of the storage device is varied up to a certain capacity, and remains constant thereafter.
This approach can be used to calculate the optimal capacity of the storage to 
be installed at the residence. We also show the learnings of the RL agent through the course of training and the strategies followed by the agent when the capacity of the storage device is varied. As part of the future work, we plan to include other parameters of the storage system like cost, lifetime, etc., which have not been incorporated in this study. The payback period of the battery can be calculated by taking these additional parameters into consideration.

\begin{acks}
We thank the members of Smart Energy Informatics Lab, IIT Bombay
for their immense support.
\end{acks}

\bibliographystyle{ACM-Reference-Format}
\bibliography{sample-base.bbl}

% \appendix









\end{document}